\documentclass[11pt,letterpaper]{article}
\usepackage{naaclhlt2016}
\usepackage{times}
\usepackage{latexsym}
\usepackage{url}
\usepackage{graphicx}
\usepackage{booktabs}

\aclfinalcopy 

\title{Hunting for Troll Comments in News Community Forums}

\author{Todor Mihaylov \\
Institute for Computational Linguistics\thanks{\ \ This research started in the Sofia University.}\\
Heidelberg University\\
Heidelberg, Germany\\
  {\tt mihaylov@cl.uni-heidelberg.de} \\\And
  Preslav Nakov \\
  Qatar Computing Research Institute\\
  Hamad bin Khalifa University\\
  P.O. box 5825, Doha, Qatar \\
  {\tt pnakov@qf.org.qa}\\}

\date{}

\begin{document}
\maketitle
\begin{abstract}
There are different definitions of what a troll is. Certainly, a troll can be somebody who teases people to make them angry, or somebody who offends people, or somebody who wants to dominate any single discussion, or somebody who tries to manipulate people's opinion (sometimes for money), etc. The last definition is the one that dominates the public discourse in Bulgaria and Eastern Europe, and this is our focus in this paper. 

In our work, we examine two types of opinion manipulation trolls: paid trolls that have been revealed from leaked ``reputation management contracts'' and ``mentioned trolls'' that have been called such by several different people.
We show that these definitions are sensible: 
we build two classifiers that can distinguish a post by such a \emph{paid troll} from one by a \emph{non-troll} with 81-82\% accuracy; the same classifier achieves 81-82\% accuracy on so called \emph{mentioned troll} vs. \emph{non-troll} posts.
\end{abstract}

\section{Introduction}
\label{sec:intro}

The practice of using Internet trolls for opinion manipulation has been reality since the rise of Internet and community forums. It has been shown that user opinions about products, companies and politics can be influenced by opinions posted by other online users in online forums and social networks 
\cite{Dellarocas06}. This makes it easy for companies and political parties to gain popularity by paying for ``reputation management'' to people that write in discussion forums and social networks fake opinions from fake profiles.

Opinion manipulation campaigns are often
launched using ``personal management software'' that allows a user to open multiple accounts and to appear like several different people.
Over time, some forum users developed sensitivity about trolls, and started publicly exposing them. 
Yet, it is hard for forum administrators to block them as trolls try formally not to violate the forum rules. 
In our work, we examine two types of opinion manipulation trolls: paid trolls that have been revealed from leaked ``reputation management contracts''\footnote{The independent Bulgarian media Bivol published a leaked contract described the following services in favor of the government:\textit{``Monthly posting online of 250 comments by virtual users with varied, typical and evolving profiles from different (non-recurring) IP addresses to inform, promote, balance or counteract. The intensity of the provided online presence will be adequately distributed and will correspond to the political situation in the country.''}
See \scriptsize \url{http://bivol.bg/en/category/b-files-en/b-files-trolls-en}} and ``mentioned trolls'' that have been called such by several different people.

\section{Related Work}
\label{sec:related}

\emph{Troll detection} was addressed semantic analysis \cite{cambria2010not} and domain-adapting sentiment analysis \cite{Seah2015}. There are also studies on general troll behavior \cite{herring2002searching,buckels2014trolls}.

\emph{Astroturfing} and misinformation have been addressed in the context of political elections using microblogging data \cite{Ratkiewicz:2011:TMS:1963192.1963301},
\emph{fake profile detection} has been studied in the context of cyber-bullying \cite{galan2014supervised}. 

A related research line is on \emph{offensive language} use \cite{xu2010filtering}.
This is related to \emph{cyber-bullying}, which has been  detected using sentiment analysis \cite{Xu:2012:FLS}, graph-based approaches over signed social networks \cite{Ortega20122884,kumar2014accurately}, and
lexico-syntactic features about user's writing style 
\cite{chen2012detecting}. 

\emph{Trustworthiness} on the Web is another relevant research direction \cite{rowe2009assessing}.
Detecting untruthful and deceptive information has been studied using both psychology and computational linguistics \cite{ott2011}. 

A related problem is \emph{Web spam detection}, which has been addressed
using spam keyword spotting \cite{dave2003mining}, lexical affinity of arbitrary words to spam content \cite{hu2004mining}, frequency of punctuation and word co-occurrence \cite{li2006combining}. 
See \cite{Castillo:2011:AWS} for an overview on adversarial web search.

In our previous work, we focused on opinion manipulation \emph{trolls} \cite{Mihaylov2015FindingOM} and on modeling the behavior of exposed vs. paid trolls \cite{Mihaylov2015ExposingPO}. Here, we go beyond user profile and we try to detect individual troll vs. non-troll \emph{comments} in a news community forum based on both text and metadata.

\begin{table}[t]
\centering
\begin{tabular}{lr}
\textbf{Object}          & \textbf{Count}   \\  \hline 
Publications       & 34,514   \\ 
Comments           & 1,930,818 \\
  -of which replies& 897,806 \\
User profiles      & 14,598   \\
Topics             & 232     \\ 
Tags               & 13,575   \\  \hline
\end{tabular}
\vspace{-1pt}
\caption{\label{table:Data:statistics}Statistics about our dataset.}
\vspace{-1pt}
\end{table}

\begin{table}[t]
\centering
\begin{tabular}{lcr}
\textbf{Label}                   & \textbf{Comments}  \\\hline
Paid troll comments           & 650    \\
Mentioned troll comments & 578        \\
Non-troll comments       & 650+578        \\
\hline
\end{tabular}
\caption{Comments selected for experiments.}
\label{table:comm-stat}
\end{table}

\section{Data}
\label{sec:data}

We crawled the largest community forum in Bulgaria, that of Dnevnik.bg, a daily newspaper (in Bulgarian) that requires users to be signed in order to read and comment. 
The platform allows users to comment on news, to reply to other users' comments and to vote on them with thumbs up/down. 
We crawled 
the \emph{Bulgaria}, \emph{Europe}, and \emph{World} categories
for the period 01-Jan-2013 to 01-Apr-2015, together with comments and user profiles:
34,514 publications on 232 topics with 13,575 tags and 1,930,818 comments (897,806 of them replies) by 14,598 users; see Table~\ref{table:Data:statistics}.
We then extracted comments by \emph{paid trolls} vs. \emph{mentioned trolls} vs. \emph{non-trolls}; see Table~\ref{table:comm-stat}.

\textbf{Paid troll comments:} 
We collected them from the leaked reputation management documents, which included 10,150 paid troll comments: 2,000 in Facebook, and 8,150 in news community forums. The latter  included 650 posted in the forum of Dnevnik.bg, which we used in our experiments.

\textbf{Mentioned troll comments:} We further collected 1,140 comments that have been replied to with an accusation of being troll comments. We considered a comment as a potential accusation if (\emph{i})~it was a reply to a comment, and (\emph{ii}) it contained words such as \emph{troll} or \emph{murzi(lka)}.\footnote{Commonly believed in Bulgaria to mean \emph{troll} in Russian (which it does not).} Two annotators checked these comments and found 578 actual accusations. The inter-annotator agreement was substantial: Cohen's Kappa of 0.82. Moreover, a simple bag-of-words classifier could find these 578 accusations with an F$_1$-score of 0.85.
Here are some examples (translated):

\begin{small}

{\bf Accusation:} \textit{``To comment from ``Prorok Ilia'': I can see that you are a red troll by the words that you are using''}

{\bf Accused troll's comment:} \textit{This Boyko\footnote{The Bulgarian Prime Minister Mr. Boyko Borisov.} is always in your mind! You only think of him. We like Boko the Potato (the favorite of the Lamb), the way we like the Karlies.}

{\bf Paid troll's comment:} \textit{in the previous protests, the entire country participated, but now we only see the paid fans of GERB.\footnote{Boyko Borisov's party GERB had fallen down due to protests and here is being accused of organizing protests in turn against the new Socialist government that replaced it.} These are not true protests, but chaotic happenings.}

\end{small}

\textbf{Non-troll comments} are those posted by users that have at least 100 comments in the forum and have never been accused of being trolls.
We selected 650 non-troll comments for the paid trolls,
and other 578 
for the mentioned trolls as follows:
for each paid or mentioned troll comment, we selected a non-troll comment at random \emph{from the same thread}. Thus, we have two separate non-troll sets of 650 and of 	578 comments.

\section{Features}
\label{sec:features}
We train a classifier to distinguish \emph{troll} (\emph{paid} or \emph{mentioned}) vs. \emph{non-troll} comments using the following features:

\textbf{Bag of words.}  We use words and their frequencies as features,
after stopword filtering.\footnote{\url{http://members.unine.ch/jacques.savoy/clef/bulgarianST.txt}}

\textbf{Bag of stems.} We further experiment with bag of stems, where we stem the words with the BulStem stemmer \cite{Nakov:2003:BIS,Nakov2003Bulstem}.

\textbf{Word $n$-grams.} We also experiment with 2- and 3-word $n$-grams.

\textbf{Char $n$-grams.} We further use character $n$-grams, where for each word token we extract all $n$ consecutive characters. We use $n$-grams of length 3 and 4 only as other values did not help.

\textbf{Word prefix.} For each word token, we extract the first 3 or 4 consecutive characters.

\textbf{Word suffix.} For each word token, we take the last 3 or 4 consecutive characters.

\textbf{Emoticons.} We extract the standard HTML-based emoticons used in the forum of Dnevnik.bg.

\textbf{Punctuation count.} 
We count the number of exclamation marks, dots, and question marks, both single and elongated, the number of words, and the number of \textsc{all caps} words.

\textbf{Metadata.} We use the time when comments were posted (worktime: 9:00-19:00h vs. night: 21:00-6:00h), part of the week (workdays: Mon-Fri vs. weekend: Sat-Sun), and the rank of the comment divided by the number of comments in the thread.

\textbf{Word2Vec clusters.} 
We trained word2vec on 80M words from 34,514 publications and 1,930,818 comments in our forum,
obtaining 268,617 word vectors, which we grouped into 5,372 clusters using K-Means clustering, and then we use these clusters as features.

\textbf{Sentiment.} We use features derived from MPQA Subjectivity Lexicon \cite{Wilson:2005:RCP:1220575.1220619} and NRC Emotion Lexicon \cite{Mohammad13} and the lexicon of \newcite{hu2004mining}. Originally these lexicons were built for English, but we translated them to Bulgarian using Google Translate.
Then, we reused the sentiment analysis pipeline from \cite{velichkov-EtAl:2014:SemEval}, which we adapted for Bulgarian.

\textbf{Bad words.} We use the number of bad words in the comment as a feature. The words come from the \textit{Bad words list v2.0}, which contains 458 bad words collected for a filter of forum or IRC channels in English.\footnote{\url{http://urbanoalvarez.es/blog/2008/04/04/bad-words-list/}}
We translated this list to Bulgarian using Google Translate and we removed duplicates to obtain \textit{Bad\_Words\_Bg\_1}. We further used the above word2vec model to find the three most similar words for each bad word in \textit{Bad\_Words\_Bg\_1}, and we constructed another lexicon: \textit{Bad\_Words\_Bg\_3}.\footnote{\label{fn:bg-lexicons}\url{http://github.com/tbmihailov/gate-lang-bulgarian-gazetteers/} - GATE resources for Bulgarian, including sentiment lexicons, bad words lexicons, politicians' names, etc.}
Finally, we generate two features: one for each lexicon.

\textbf{Mentions.} We noted that trolls use diminutive names or humiliating nicknames when referring to politicians that they do not like, but use full or family names for people that they respect. Based on these observations, we constructed several lexicons with Bulgarian politician names, their variations and nicknames (see footnote \ref{fn:bg-lexicons}), and we generated a mention count feature for each lexicon.

\textbf{POS tag distribution.} We also use features based on part of speech (POS). We tag using GATE \cite{Cunningham:2011a}
with a simplified model trained on a transformed version of the BulTreeBank-DP \cite{Simov02buildinga}. For each POS tag type, we take the number of occurrences in the text divided by the total number of tokens.
We use both fine-grained and course-grained POS tags, e.g., from the POS tag \textit{Npmsi}, we generate three tags: \textit{Npmsi}, \textit{N} and \textit{Np}.

\textbf{Named entities}. We also use the occurrence of named entities as features. For extracting named entities such as \textit{location}, \textit{country}, \textit{person\_name}, \textit{date\_unit}, etc., we use the lexicons that come with Gate's ANNIE \cite{Cunningham2002} pipeline, which we translated to Bulgarian.
In future work, we plan to use a better named entity recognizer based on CRF \cite{georgiev-EtAl:2009:RANLP09}.

\section{Experiments and Evaluation}
\label{sec:experiments}

\begin{table}[t]
\centering
\begin{tabular}{lll}
\textbf{Features}  & \textbf{F} & \textbf{Acc}  \\\hline
All $-$ char n-grams & 79.24 & 78.54\\
All $-$ word suff & 78.58 & 78.20\\
All $-$ word preff & 78.51 & 78.02\\
All $-$ bow stems & 78.32 & 77.85\\
All $-$ bow with stop & 78.25 & 77.77\\
All $-$ bad words & 78.10 & 77.68\\
All $-$ emoticons & 78.08 & 77.76\\
All $-$ mentions & 78.06 & 77.68\\
All & 78.06 & 77.68\\
All $-$ (bow, no stop) & 78.04 & 77.68\\
All $-$ NE & 77.98 & 77.59\\
All $-$ sentiment & 77.95 & 77.51\\
All $-$ POS & 77.80 & 77.33\\
All $-$ w2v clusters & 77.79 & 77.25\\
All $-$ word 3-grams & 77.69 & 77.33\\
All $-$ word 2-grams & 77.62 & 77.25\\
All $-$ punct & 77.29 & 76.90\\
All $-$ metadata & 70.77 & 70.94\\\hline
Baseline   & 50.00 & 50.00\\\hline
\end{tabular}
\caption{\textbf{Mentioned troll vs. non-troll comments.} Ablation excluding feature groups.}
\label{table:mentioned-vs-non-allfeatures}
\end{table}

\begin{table}[t]
\centering
\begin{tabular}{lll}
\textbf{Features}  & \textbf{F} & \textbf{Acc}  \\\hline
All $-$ char n-grams & 81.08 & 81.77\\
All $-$ word suff & 81.00 & 81.77\\
All $-$ word preff & 80.83 & 81.62\\
All $-$ bow with stop & 80.67 & 81.54\\
All $-$ sentiment & 80.63 & 81.46\\
All $-$ word 2-grams & 80.62 & 81.46\\
All $-$ w2v clusters & 80.54 & 81.38\\
All $-$ word 3-grams & 80.46 & 81.38\\
All $-$ punct & 80.40 & 81.23\\
All $-$ mentions & 80.40 & 81.31\\
All & 80.40 & 81.31\\
All $-$ bow stems & 80.37 & 81.31\\
All $-$ emoticons & 80.33 & 81.15\\
All $-$ bad words & 80.09 & 81.00\\
All $-$ NE & 80.00 & 80.92\\
All $-$ POS & 79.77 & 80.69\\
All $-$ (bow, no stop) & 79.46 & 80.38\\
All $-$ metadata & 75.37 & 76.62\\\hline
Baseline   & 50.00 & 50.00\\\hline
\end{tabular}
\caption{\textbf{Paid troll vs. non-troll comments.} Ablation excluding feature groups.}
\label{table:paid-vs-non-allfeatures}
\end{table}

We train an L2-regularized Logistic Regression with LIBLINEAR \cite{Fan2008Liblinear} as implemented in \textsc{scikit-learn} \cite{scikit-learn}, 
using scaled and normalized features.
As we have perfectly balanced sets of 650 positive and 650 negative examples for \textit{paid troll} vs. \textit{non-trolls} and 578 positive and 578 negative examples for \textit{mentioned troll} vs. \textit{non-trolls}, the baseline accuracy is 50\%.
Below, we report F-score and accuracy with cross-validation.

Table~\ref{table:mentioned-vs-non-allfeatures}, shows the results for experiments to distinguish comments by mentioned trolls vs. such by non-trolls,
using all features, as well as when excluding individual feature groups.
We can see that excluding character $n$-grams, word suffixes and word prefixes from the features, as well as excluding bag of words with stems or stop words, yields performance gains; the most sizable gain is when excluding char $n$-grams, which yields one point of improvement. Excluding bad words usage and emoticons also improves the performance but insignificantly, which might be because they are covered by the bag of words features. 

Excluding any of the other features hurts performance, the two most important features to keep being metadata (as it allows us to see the time of posting), and bag of words without stopwords (which looks at the vocabulary choice that mentioned trolls use differently from regular users).

Table~\ref{table:paid-vs-non-allfeatures} shows the results for telling apart comments by paid trolls vs. such by non-trolls, using cross-validation and ablation with the same features as for the mentioned trolls. There are several interesting observations we can make. First, we can see that the overall accuracy for finding paid trolls is slightly higher, namely 81.02, vs. 79.24 for mentioned trolls. The most helpful feature again is metadata, but this time it is less helpful (excluding it yields a drop of 5 points vs. 8 points before). The least helpful feature again are character $n$-grams. The remaining features fall in between, and most of them yield better performance when excluded, which suggests that there is a lot of redundancy in the features.

Next, we look at individual feature groups. Table~\ref{table:mentioned-vs-non-singlegroups} shows the results for comments by mentioned trolls vs. such by non-trolls. We can see that the metadata features are by far the most important: using them alone outperforms the results when using all features by 3.5 points.

The reason could be that most troll comments are replies to other comments, while those by non-trolls are mostly not replies. Adding other features such as sentiment-based features, bad words, POS, and punctuation hurts the performance significantly. Features such as bad words are at the very bottom: they do not apply to all comments and thus are of little use alone; similarly for mentions and sentiment features, which are also quite weak in isolation. These results suggest that mentioned trolls are not that different from non-trolls in terms of language use, but have mainly different behavior in terms of replying to other users.

Table~\ref{table:paid-vs-non-singlegroups} shows a bit different picture for comments by paid trolls vs. such by non-trolls. The biggest difference is that metadata features are not so useful. Also, the strongest feature set is the combination of sentiment, bad words distribution, POS, metadata, and punctuation. This suggests that paid trolls are smart to post during time intervals and days of the week as non-trolls, but they use comments with slightly different sentiment and bad word use than non-trolls. Features based on words are also very helpful because paid trolls have to defend pre-specified key points, which limits their vocabulary use, while non-trolls are free to express themselves as they wish.

\begin{table}[t]
\centering
\begin{tabular}{lll}
\textbf{Features} & \textbf{F}   & \textbf{Acc}  \\\hline
All & 78.06 & 77.68\\\hline
Only metadata & 84.14 & 81.14\\
Sent,bad,pos,NE,meta,punct & 77.79 & 76.73\\
Only bow, no stop & 73.41 & 73.79\\
Only bow with stop & 73.41 & 73.44\\
Only bow stems & 72.43 & 72.49\\
Only word preff & 71.11 & 71.62\\
Only w2v clusters & 69.85 & 70.50\\
Only word suff & 69.17 & 68.95\\
Only word 2-grams & 68.96 & 69.29\\
Only char n-grams & 68.44 & 68.94\\
Only word 3-grams & 64.74 & 67.21\\
Only POS & 64.60 & 65.31\\
Sent,bad,pos,NE & 63.68 & 64.10\\
Only sent,bad & 63.66 & 64.44\\
Only emoticons & 63.30 & 64.96\\
Sent,bad,ment,NE & 63.11 & 64.01\\
Only punct & 63.09 & 64.79\\
Only sentiment & 62.50 & 63.66\\
Only NE & 62.45 & 64.27\\
Only mentions & 62.41 & 64.10\\
Only bad words & 62.27 & 64.01\\\hline
Baseline                   & 50.00   & 50.00\\\hline
\end{tabular}
\caption{\textbf{Mentioned troll comments vs. non-troll comments.} Results for individual feature groups.}
\label{table:mentioned-vs-non-singlegroups}
\end{table}

\section{Discussion}
\label{sec:discuss}

Overall, we have seen that our classifier for telling apart comments by mentioned trolls vs. such by non-trolls performs almost equally well for paid trolls vs. non-trolls, where the non-troll comments are sampled from the same threads that the 
troll comments come from. Moreover, the most and the least important features ablated from all are also similar. This suggests that mentioned trolls are very similar to paid trolls (except for their reply rate, time and day of posting patterns).

\begin{table}[t]
\centering
\begin{tabular}{lll}
\textbf{Features} & \textbf{F} & \textbf{Acc}  \\\hline
All & 80.40 & 81.31\\\hline
Sent,bad,pos,NE,meta,punct & 78.04 & 78.15 \\
Only bow, no stop & 75.95 & 76.46 \\
Only word 2-grams & 75.55 & 74.92 \\
Only bow with stop & 75.27 & 75.62 \\
Only bow stems & 75.25 & 76.08 \\
Only w2v clusters & 74.20 & 74.00 \\
Only word preff & 74.01 & 74.77 \\
Sent,bad,pos,NE & 73.89 & 73.85 \\
Only metadata & 73.79 & 72.54 \\
Only char n-grams & 73.02 & 74.23 \\
Only POS & 72.94 & 72.69 \\
Only word suff & 72.03 & 72.69 \\
Only word 3-grams & 69.20 & 68.00 \\
Only punct & 66.80 & 65.00 \\
Only NE & 66.54 & 64.77 \\
Sent,bad,ment,NE & 66.04 & 64.92 \\
Only sentiment & 64.28 & 62.62 \\
Only mentions & 63.28 & 61.46 \\
Only sent,bad & 63.14 & 61.54 \\
Only emoticons & 62.95 & 61.00 \\
Only bad words & 62.22 & 60.85 \\\hline
Baseline            & 50.00  & 50.00\\\hline

\end{tabular}
\caption{\textbf{Paid troll vs. non-troll comments.} Results for individual feature groups.}
\label{table:paid-vs-non-singlegroups}
\end{table}

\begin{table}[tbh]
\centering

\begin{tabular}{llllll}
    & \textbf{5}     & \textbf{10}  & \textbf{15}    & \textbf{20}    \\\hline
Acc  & 80.70 & 81.08 & 83.41 & 85.59 \\
Diff & +8.46  & +18.51 & +30.81 & +32.26\\\hline
\end{tabular}
\caption{\textbf{Mentioned troll vs. non-troll \emph{users (not comments!)}}. Experiments with different number of minimum mentions for January, 2015. 
`Diff'' is the difference from the majority class baseline.}
\label{table:2class-comm-diff-mentions}
\end{table}

However, using just mentions might be a ``witch hunt'': some users could have been accused of being ``trolls'' unfairly. 
One way to test this is to look not at comments, but at users and to see which users were called trolls by several different other users.
Table \ref{table:2class-comm-diff-mentions} shows the results for distinguishing users with a given number of alleged troll comments from non-troll users; the classification is based on all comments by the corresponding users. We can see that finding users who have been called trolls more often is easier, which suggests they might be trolls indeed.

\section{Conclusion and Future Work}
\label{sec:future}

We have presented experiments in predicting whether a comment is written by a troll or not, where we define troll as somebody who was called such by other people. We have shown that this is a useful definition and that comments by mentioned trolls are similar to such by confirmed paid trolls.

\textbf{Acknowledgments.} 
This research is part of the Interactive sYstems for Answer Search (Iyas) project, which is developed by the Arabic Language Technologies (ALT) group at the Qatar Computing Research Institute (QCRI), Hamad bin Khalifa University (HBKU), part of Qatar Foundation in collaboration with MIT-CSAIL.

\bibliographystyle{naaclhlt2016}
\bibliography{bib}

\begin{thebibliography}{}

\bibitem[\protect\citename{Buckels \bgroup et al.\egroup
  }2014]{buckels2014trolls}
Erin~E Buckels, Paul~D Trapnell, and Delroy~L Paulhus.
\newblock 2014.
\newblock Trolls just want to have fun.
\newblock {\em Personality and individual Differences}, 67:97--102.

\bibitem[\protect\citename{Cambria \bgroup et al.\egroup }2010]{cambria2010not}
Erik Cambria, Praphul Chandra, Avinash Sharma, and Amir Hussain.
\newblock 2010.
\newblock Do not feel the trolls.
\newblock In {\em Proceedings of the 3rd International Workshop on Social Data
  on the Web}, SDoW~'10, Shanghai, China.

\bibitem[\protect\citename{Castillo and Davison}2011]{Castillo:2011:AWS}
Carlos Castillo and Brian~D. Davison.
\newblock 2011.
\newblock Adversarial web search.
\newblock {\em Found. Trends Inf. Retr.}, 4(5):377--486, May.

\bibitem[\protect\citename{Chen \bgroup et al.\egroup }2012]{chen2012detecting}
Ying Chen, Yilu Zhou, Sencun Zhu, and Heng Xu.
\newblock 2012.
\newblock Detecting offensive language in social media to protect adolescent
  online safety.
\newblock In {\em Proceedings of the 2012 International Conference on Privacy,
  Security, Risk and Trust and of the 2012 International Conference on Social
  Computing}, PASSAT/SocialCom~'12, pages 71--80, Amsterdam, Netherlands.

\bibitem[\protect\citename{Cunningham \bgroup et al.\egroup
  }2002]{Cunningham2002}
Hamish Cunningham, Diana Maynard, Kalina Bontcheva, and Valentin Tablan.
\newblock 2002.
\newblock {GATE}: an architecture for development of robust {HLT} applications.
\newblock In {\em Proceedings of 40th Annual Meeting of the Association for
  Computational Linguistics}, ACL~'02, pages 168--175, Philadelphia,
  Pennsylvania, USA.

\bibitem[\protect\citename{Cunningham \bgroup et al.\egroup
  }2011]{Cunningham:2011a}
Hamish Cunningham, Diana Maynard, and Kalina Bontcheva.
\newblock 2011.
\newblock {\em Text Processing with {GATE}}.
\newblock Gateway Press CA.

\bibitem[\protect\citename{Dave \bgroup et al.\egroup }2003]{dave2003mining}
Kushal Dave, Steve Lawrence, and David~M Pennock.
\newblock 2003.
\newblock Mining the peanut gallery: Opinion extraction and semantic
  classification of product reviews.
\newblock In {\em Proceedings of the 12th International World Wide Web
  conference}, WWW~'03, pages 519--528, Budapest, Hungary.

\bibitem[\protect\citename{Dellarocas}2006]{Dellarocas06}
Chrysanthos Dellarocas.
\newblock 2006.
\newblock Strategic manipulation of internet opinion forums: Implications for
  consumers and firms.
\newblock {\em Management Science}, 52(10):1577--1593.

\bibitem[\protect\citename{Fan \bgroup et al.\egroup }2008]{Fan2008Liblinear}
Rong-En Fan, Kai-Wei Chang, Cho-Jui Hsieh, Xiang-Rui Wang, and Chih-Jen Lin.
\newblock 2008.
\newblock Liblinear: A library for large linear classification.
\newblock {\em J. Mach. Learn. Res.}, 9:1871--1874, June.

\bibitem[\protect\citename{Gal{\'a}n-Garc{\'\i}a \bgroup et al.\egroup
  }2014]{galan2014supervised}
Patxi Gal{\'a}n-Garc{\'\i}a, Jos{\'e}~Gaviria de~la Puerta, Carlos~Laorden
  G{\'o}mez, Igor Santos, and Pablo~Garc{\'\i}a Bringas.
\newblock 2014.
\newblock Supervised machine learning for the detection of troll profiles in
  {T}witter social network: Application to a real case of cyberbullying.
\newblock In {\em Proceedings of the International Joint Conference
  SOCO’13-CISIS’13-ICEUTE’13}, Advances in Intelligent Systems and
  Computing, pages 419--428. Springer International Publishing.

\bibitem[\protect\citename{Georgiev \bgroup et al.\egroup
  }2009]{georgiev-EtAl:2009:RANLP09}
Georgi Georgiev, Preslav Nakov, Kuzman Ganchev, Petya Osenova, and Kiril Simov.
\newblock 2009.
\newblock Feature-rich named entity recognition for {B}ulgarian using
  conditional random fields.
\newblock In {\em Proceedings of the International Conference Recent Advances
  in Natural Language Processing}, RANLP~'09, pages 113--117, Borovets,
  Bulgaria.

\bibitem[\protect\citename{Herring \bgroup et al.\egroup
  }2002]{herring2002searching}
Susan Herring, Kirk Job-Sluder, Rebecca Scheckler, and Sasha Barab.
\newblock 2002.
\newblock Searching for safety online: Managing ``trolling'' in a feminist
  forum.
\newblock {\em The Information Society}, 18(5):371--384.

\bibitem[\protect\citename{Hu and Liu}2004]{hu2004mining}
Minqing Hu and Bing Liu.
\newblock 2004.
\newblock Mining and summarizing customer reviews.
\newblock In {\em Proceedings of the 10th ACM SIGKDD International Conference
  on Knowledge Discovery and Data Mining}, KDD~'04, pages 168--177, Seattle,
  Washington, USA.

\bibitem[\protect\citename{Kumar \bgroup et al.\egroup
  }2014]{kumar2014accurately}
Srijan Kumar, Francesca Spezzano, and VS~Subrahmanian.
\newblock 2014.
\newblock Accurately detecting trolls in slashdot zoo via decluttering.
\newblock In {\em Proceedings of the 2014 IEEE/ACM International Conference on
  Advances in Social Network Analysis and Mining}, ASONAM~'14, pages 188--195,
  Beijing, China.

\bibitem[\protect\citename{Li \bgroup et al.\egroup }2006]{li2006combining}
Wenbin Li, Ning Zhong, and Chunnian Liu.
\newblock 2006.
\newblock Combining multiple email filters based on multivariate statistical
  analysis.
\newblock In {\em Foundations of Intelligent Systems}, pages 729--738.
  Springer.

\bibitem[\protect\citename{Mihaylov \bgroup et al.\egroup
  }2015a]{Mihaylov2015FindingOM}
Todor Mihaylov, Georgi Georgiev, and Preslav Nakov.
\newblock 2015a.
\newblock Finding opinion manipulation trolls in news community forums.
\newblock In {\em Proceedings of the Nineteenth Conference on Computational
  Natural Language Learning}, CoNLL~'15, pages 310--314, Beijing, China.

\bibitem[\protect\citename{Mihaylov \bgroup et al.\egroup
  }2015b]{Mihaylov2015ExposingPO}
Todor Mihaylov, Ivan Koychev, Georgi Georgiev, and Preslav Nakov.
\newblock 2015b.
\newblock Exposing paid opinion manipulation trolls.
\newblock In {\em Proceedings of the International Conference Recent Advances
  in Natural Language Processing}, RANLP~'15, pages 443--450, Hissar, Bulgaria.

\bibitem[\protect\citename{Mohammad and Turney}2013]{Mohammad13}
Saif~M. Mohammad and Peter~D. Turney.
\newblock 2013.
\newblock Crowdsourcing a word-emotion association lexicon.
\newblock {\em Computational Intelligence}, 29(3):436--465.

\bibitem[\protect\citename{Nakov}2003a]{Nakov:2003:BIS}
Preslav Nakov.
\newblock 2003a.
\newblock Building an inflectional stemmer for {B}ulgarian.
\newblock In {\em Proceedings of the 4th International Conference on Computer
  Systems and Technologies: E-Learning}, CompSysTech '03, pages 419--424,
  Rousse, Bulgaria.

\bibitem[\protect\citename{Nakov}2003b]{Nakov2003Bulstem}
Preslav Nakov.
\newblock 2003b.
\newblock {BulStem}: Design and evaluation of inflectional stemmer for
  {B}ulgarian.
\newblock In {\em Proceedings of Workshop on Balkan Language Resources and
  Tools (1st Balkan Conference in Informatics), Thessaloniki, Greece, November,
  2003}.

\bibitem[\protect\citename{Ortega \bgroup et al.\egroup }2012]{Ortega20122884}
F.~Javier Ortega, José~A. Troyano, Fermín~L. Cruz, Carlos~G. Vallejo, and
  Fernando Enríquez.
\newblock 2012.
\newblock Propagation of trust and distrust for the detection of trolls in a
  social network.
\newblock {\em Computer Networks}, 56(12):2884 -- 2895.

\bibitem[\protect\citename{Ott \bgroup et al.\egroup }2011]{ott2011}
Myle Ott, Yejin Choi, Claire Cardie, and Jeffrey~T. Hancock.
\newblock 2011.
\newblock Finding deceptive opinion spam by any stretch of the imagination.
\newblock In {\em Proceedings of the 49th Annual Meeting of the Association for
  Computational Linguistics: Human Language Technologies - Volume 1}, HLT '11,
  pages 309--319, Portland, Oregon.

\bibitem[\protect\citename{Pedregosa \bgroup et al.\egroup }2011]{scikit-learn}
F.~Pedregosa, G.~Varoquaux, A.~Gramfort, V.~Michel, B.~Thirion, O.~Grisel,
  M.~Blondel, P.~Prettenhofer, R.~Weiss, V.~Dubourg, J.~Vanderplas, A.~Passos,
  D.~Cournapeau, M.~Brucher, M.~Perrot, and E.~Duchesnay.
\newblock 2011.
\newblock Scikit-learn: Machine learning in {P}ython.
\newblock {\em Journal of Machine Learning Research}, 12:2825--2830.

\bibitem[\protect\citename{Ratkiewicz \bgroup et al.\egroup
  }2011]{Ratkiewicz:2011:TMS:1963192.1963301}
Jacob Ratkiewicz, Michael Conover, Mark Meiss, Bruno Gon\c{c}alves, Snehal
  Patil, Alessandro Flammini, and Filippo Menczer.
\newblock 2011.
\newblock Truthy: Mapping the spread of astroturf in microblog streams.
\newblock In {\em Proceedings of the 20th International Conference Companion on
  World Wide Web}, WWW '11, pages 249--252, Hyderabad, India.

\bibitem[\protect\citename{Rowe and Butters}2009]{rowe2009assessing}
Matthew Rowe and Jonathan Butters.
\newblock 2009.
\newblock {Assessing Trust: Contextual Accountability}.
\newblock In {\em Proceedings of the First Workshop on Trust and Privacy on the
  Social and Semantic Web}, SPOT~'09, Heraklion, Greece.

\bibitem[\protect\citename{Seah \bgroup et al.\egroup }2015]{Seah2015}
Chun-Wei Seah, Hai~Leong Chieu, Kian Ming~Adam Chai, Loo-Nin Teow, and Lee~Wei
  Yeong.
\newblock 2015.
\newblock Troll detection by domain-adapting sentiment analysis.
\newblock In {\em Proceedings of the 18th International Conference on
  Information Fusion}, FUSION~'15, pages 792--799, Washington, DC, USA.

\bibitem[\protect\citename{Simov \bgroup et al.\egroup }2002]{Simov02buildinga}
Kiril Simov, Petya Osenova, Milena Slavcheva, Sia Kolkovska, Elisaveta
  Balabanova, Dimitar Doikoff, Krassimira Ivanova, Er~Simov, and Milen
  Kouylekov.
\newblock 2002.
\newblock Building a linguistically interpreted corpus of {B}ulgarian: the
  {BulTreeBank}.
\newblock In {\em Proceedings of the Third International Conference on Language
  Resources and Evaluation}, LREC~'02, Canary Islands, Spain.

\bibitem[\protect\citename{Velichkov \bgroup et al.\egroup
  }2014]{velichkov-EtAl:2014:SemEval}
Boris Velichkov, Borislav Kapukaranov, Ivan Grozev, Jeni Karanesheva, Todor
  Mihaylov, Yasen Kiprov, Preslav Nakov, Ivan Koychev, and Georgi Georgiev.
\newblock 2014.
\newblock {SU-FMI}: System description for {SemEval}-2014 task 9 on sentiment
  analysis in {T}witter.
\newblock In {\em Proceedings of the 8th International Workshop on Semantic
  Evaluation}, SemEval~'14, pages 590--595, Dublin, Ireland.

\bibitem[\protect\citename{Wilson \bgroup et al.\egroup
  }2005]{Wilson:2005:RCP:1220575.1220619}
Theresa Wilson, Janyce Wiebe, and Paul Hoffmann.
\newblock 2005.
\newblock Recognizing contextual polarity in phrase-level sentiment analysis.
\newblock In {\em Proceedings of the Conference on Human Language Technology
  and Empirical Methods in Natural Language Processing}, HLT '05, pages
  347--354, Vancouver, British Columbia, Canada.

\bibitem[\protect\citename{Xu and Zhu}2010]{xu2010filtering}
Zhi Xu and Sencun Zhu.
\newblock 2010.
\newblock Filtering offensive language in online communities using grammatical
  relations.
\newblock In {\em Proceedings of the Seventh Annual Collaboration, Electronic
  Messaging, Anti-Abuse and Spam Conference}, CEAS~'10, Redmond, Washington,
  USA.

\bibitem[\protect\citename{Xu \bgroup et al.\egroup }2012]{Xu:2012:FLS}
Jun-Ming Xu, Xiaojin Zhu, and Amy Bellmore.
\newblock 2012.
\newblock Fast learning for sentiment analysis on bullying.
\newblock In {\em Proceedings of the First International Workshop on Issues of
  Sentiment Discovery and Opinion Mining}, WISDOM '12, pages 10:1--10:6,
  Beijing, China.

\end{thebibliography}

\end{document}